\documentclass{article}

\usepackage{arxiv}

\usepackage[utf8]{inputenc} 
\usepackage[T1]{fontenc}    
\usepackage{hyperref}       
\usepackage{url}            
\usepackage{booktabs}       
\usepackage{amsfonts}       
\usepackage{nicefrac}       
\usepackage{microtype}      
\usepackage{graphicx}
\graphicspath{ {./pics/} }
\usepackage{amsmath,bm}
\usepackage[noadjust]{cite}
\usepackage{adjustbox}

\usepackage{hyperref}      
\usepackage{caption} 
\usepackage{subcaption}
\usepackage{tabularx}
\usepackage{bbm}
\usepackage{multirow} 
\usepackage[linesnumbered,ruled,vlined]{algorithm2e} 

\DeclareCaptionLabelFormat{tableandfigure}{#1~#2  \&  \tablename~\thetable}

\title{S\&P 500 Stock Price Prediction Using Technical, Fundamental and Text Data}

\author{
  Shan Zhong\\
  Department of Statistics\\
  University of South Carolina\\
  Columbia, SC 29208\\
  \texttt{zhongs@email.sc.edu} \\
	\And
  David B.\ Hitchcock\\
  Department of Statistics\\
  University of South Carolina\\
  Columbia, SC 29208\\
  \texttt{hitchcock@stat.sc.edu }
}

\begin{document}
\maketitle

\begin{abstract}
We summarized both common and novel predictive models used for stock price prediction and combined them with technical indices, fundamental characteristics and text-based sentiment data to predict S\&P stock prices. A 66.18\% accuracy in S\&P 500 index directional prediction and 62.09\% accuracy in individual stock directional prediction was achieved by combining different machine learning models such as Random Forest and LSTM together into state-of-the-art ensemble models. The data we use contains weekly historical prices, finance reports, and text information from news items associated with 518 different common stocks issued by current and former S\&P 500 large-cap companies, from January 1, 2000 to December 31, 2019. Our study's innovation includes utilizing deep language models to categorize and infer financial news item sentiment; fusing different models containing different combinations of variables and stocks to jointly make predictions; and overcoming the insufficient data problem for machine learning models in time series by using data across different stocks.\\
\end{abstract}



\section{INTRODUCTION AND BACKGROUND KNOWLEDGE}

Modeling and predicting future stock prices have long been important but difficult problems for researchers \cite{mandelbrot1997variation}\cite{samuelson2016proof}. Accurate prediction of future stock prices provides a foundation for the risk analysis and investment decision making that are necessary in stock trading. On the other hand, recent studies \cite{qian2007stock}\cite{bollen2011twitter} have shown stock markets are predictable with more than 50 percent accuracy, which is a breakthrough that may potentially belie the statement in the Efficient Market Hypothesis (EMH)\cite{fama1970efficient} that stock prices are unpredictable. Then the questions remaining are how much accuracy can we achieve with state-of-the-art models, and how can we integrate these new predictive models together with different kinds of data. Surveys \cite{fischer2018reinforcement}\cite{obthong2020survey} of recent studies showed that many new methods have been proposed for prediction of stock prices, but few approaches have combined and integrated them together into one composite system efficiently, from a statistical point of view.\\

There are two popular investment strategies. One is trend following: The trend following strategies focus more on the trend movements of individual stocks and markets. The most basic trend following strategy assumes a time series with a type of momentum that suggests that people buy those stocks with recent positive returns and sell those stocks with recent negative returns \cite{hurst2017century}. The other popular investment strategy is value investing: Value investing focuses more on the intrinsic value of the companies and market. One value investing example would be buying those stocks with a high book-to-market ratio, which are the companies with high equity (company asset minus debt) to market value ratio \cite{piotroski2000value}. In our study, we combined the strengths of the ideas and the data associated with these two investment strategies, integrating them into one model.\\ 

Traditionally according to the EMH, stock prices reflect all past information, since new information is unpredictable and the stock prices will be adjusted immediately after information is made public, implying that the stock price follows a random walk \cite{fama1965behavior}. Thus the stock prices themselves and the trading volume are enough to predict future price movements. Based on assuming the stock price follows a random walk, or more generally, a stochastic process, analysts usually use traditional time series modeling techniques such as the Autoregressive Integrated Moving Average (ARIMA) model to help predict stock price. The ARIMA model has its strength in the robustness and efficiency in terms of short-run forecasting \cite{adebiyi2014comparison}\cite{ariyo2014stock}\cite{meyler1998forecasting}. However, most ARIMA models are univariate, and the data involved in these ARIMA models are usually only the historical time series stock price itself \cite{lux2007forecasting}\cite{adebiyi2014comparison}. For example, if we want to do daily forecasting, we lose the daily high, low, open, close and volume information. That means that we are unable to utilize the technical indicators which are derived directly from this stock price information that many traders use to predict the trend from a supply-demand perspective. At the same time, with machine learning models becoming popular, some researchers utilize the flexibility of machine learning algorithms to incorporate technical indicators into their model. For example, one can generate a list of related technical indicators to feed into a Random Forest (RF) or a Long Short-Term Memory (LSTM) model for directional prediction \cite{nelson2017stock}\cite{jiang2020improved}. Nevertheless, when using these machine learning models which need more parameters to fit, usually there is not enough data available, as the available data is limited based on the chosen time period. Our model resolves this issue by the following techniques: carefully selecting the corresponding predictor variables to control the number of input variables; training the machine learning algorithm across different stocks simultaneously then fine-tuning on each individual stock; creating an ensemble of models trained with different groups of stocks; and allowing multiple input and output periods for the LSTM model for more recursive training.\\

On the other hand, since the EMH was introduced, some researchers \cite{butler1992efficiency}\cite{kavussanos2001multivariate}\cite{qian2007stock} have disputed the notion and suggested that the market is inefficient and the stock prices actually do not follow random walks exactly, but rather approximate random walks. It is possible that a stock is mispriced by the market in the short run but reverts to its ``correct'' price eventually. Therefore they look at fundamental characteristics such as the price-earning ratio to check whether a stock is underpriced. There is no standard or commonly accepted baseline about what model or variables we should use for a value-investing approach. Researchers use a variety of different variables such as cash flow, interest rates, etc., in regression or empirical models \cite{yao2005dynamic}\cite{park2003empirical}\cite{atsalakis2010surveying}. In our study,  to predict the intrinsic value of stock prices, we utilized some ratios suggested by both empirical rules and from the quarterly reports that public companies must produce based on generally accepted accounting principles (GAAP) and included them into our ensemble model.\\

Besides numerical information, since 2000, the growth of the internet has greatly sped up the spread of textual information. However, for large sets of stock price data, little research has incorporated jointly into one model text data related to news items together with other factors that may affect future stock prices. In our dataset which contains over 500 stocks and 3.2 million news items, we link the stocks to the related entities mentioned in the news. The method we propose is to first recognize the news as related to each company involved, and then calculate a sentiment score of the news which can measure the negative or positive impact on the stock. As these steps can be done independently for each article, this can be accomplished efficiently. We used the Longest Common Subsequence \cite{hirschberg1977algorithms} distance metric and word embedding \cite{mikolov2013distributed} techniques to find the entities related to each stock that were mentioned in the news articles. Then a pretrained Bidirectional Encoder Representations from Transformers (BERT)\cite{devlin2018bert} model with finance domain knowledge \cite{araci2019finbert} was used to calculate the sentiment of each article.\\


\section{Literature review}

There are several different studies with the goal of predicting the S\&P 500 index and prices of stocks in the S\&P 500 index, as summarized in Table~\ref{summary_of_literature_review}. Jiang et al.\cite{jiang2020improved} utilized an ensemble stacking with logistic regression to predict stock price direction after one month from 2012 to 2019. They used a list of 24 technical and macroeconomic variables together with eight different models to do weighted shrinkage. Yu and Yan\cite{yu2020stock} used a Long Short-Term Memory (LSTM) model on daily S\&P closing prices along with wavelet smoothing to predict the next day's closing prices. They reconstructed the LSTM series to allow for multiple outputs to help train the model. Both Jiang et al. and Yu et al. utilized the LSTM model in their research, however, the effectiveness of their LSTM model was limited by the insufficient number of observations: 4,054 and 2,518 observations compared with over 300,000 and 12,000 parameters needed to fit. Ding et al.\cite{ding2015deep} utilized text embedding with convolution to predict daily S\&P 500 closing for a 10-month period in 2013. Their research showed the effect of news information gradually weakened over time, and that using news to predict daily data performs better than for weekly or monthly data, although news still has an effect in models for weekly and monthly data. Gorenc Novak and Velušček\cite{gorenc2016prediction} used technical indicators with a linear Support Vector Machine (SVM) to predict the daily high for 370 S\&P 500 companies from 2004 to 2013. They chose a 500-day rolling window period to train their model as a compromise between a smaller stationary set and a larger but less stationary set. They also did a grid search for the best model refresh period with options for 5, 10, 20, 40, 80 and 160 days, with 20 days yielding the highest accuracy. Across these distinct approaches, differences in the choice of period, preprocessing steps, and dependent variables make comparisons between different methods hard. \\

\begin{minipage}{1\linewidth}\centering
\captionof{table}{Summary of different methods}\label{summary_of_literature_review}
\begin{adjustbox}{width=\columnwidth,center}
\begin{tabular}{|l|l|l|l|c|l|}
\hline
Author             & Period                         & Variables Used                                                                    & Method                                                                     & \multicolumn{1}{l|}{\begin{tabular}[c]{@{}l@{}}Directional \\ Accuracy\end{tabular}} & Predicting                                                                                    \\ \hline
Jiang et al.\cite{jiang2020improved} & \multicolumn{1}{c|}{2012-2019} & \begin{tabular}[c]{@{}l@{}}Technical and \\ Macroeconomic indicators\end{tabular} & Ensemble stacking                                                          & 69.17\%                                                                              & \begin{tabular}[c]{@{}l@{}}One month forward \\ direction of S\&P 500 \\ closing\end{tabular} \\ \hline
Yu and Yan\cite{yu2020stock}     & \multicolumn{1}{c|}{2010-2017} & Stock Price itself                                                                & LSTM                                                                       & 58.07\%                                                                              & \begin{tabular}[c]{@{}l@{}}Daily S\&P 500 closing\\ with wavelet smoothing\end{tabular}       \\ \hline
Ding et al.\cite{ding2015deep}   & 2013                           & Text News                                                                         & \begin{tabular}[c]{@{}l@{}}Word Embedding \\ with convolution\end{tabular} & 64.21\%                                                                              & Daily S\&P 500 closing                                                                        \\ \hline
Gorenc Novak and Velušček\cite{gorenc2016prediction}  & 2004-2013                      & Technical indicators                                                              & Linear SVM                                                                 & 61.16\%                                                                              & \begin{tabular}[c]{@{}l@{}}Daily high for 370\\ S\&P 500 companies\end{tabular}               \\ \hline
\end{tabular}
\end{adjustbox}
\end{minipage}\\

One important aspect of our research is the use in our model of text data from news items. Ding et al.\cite{ding2015deep} categorized news into long-term, mid-term, and short-term events based on events that happened one month, one week, and one day ago, respectively. Then they reduced the news item into action and actors, converted the text into word embedding, and fed them into convolution neural networks for direct inference about upward or downward movements. Akita et al.\cite{akita2016deep} used textual information from news headlines to attain inference about future stock prices. They first converted news headlines to a numerical vector representation. Then they concatenated historical price information and news headline vectors for correlated groups of companies. The combined vectors were fed into one LSTM layer with 20 nodes. The predicted price for these correlated companies were then attained by regression on these 20 nodes. Bollen et al.\cite{bollen2011twitter} extracted public sentiment from social media texts, without directly inferring the meaning of these texts, to analyze public mood. They showed sentiment can help explain investor behavior and eventually help to predict the Dow Jones Industrial Average (DJIA). Following these research ideas, we also utilized the predictive power of textual information and combined it with numerical data in our hybrid model. Distinctions between our approach and others' include that Bollen et al. and Akita et al. used short phrases to fit word embedding models to get inference from text, while we used the BERT\cite{devlin2018bert} model to get accurate inference from long paragraphs of words, and we incorporated other technical and fundamental variables into the model, whereas Bollen et al. and Ding et al. only used news as inputs.\\


\section{Data Selection and Preprocessing}\label{Data selection and Preprocessing Section}

\subsection{Stock Price Data}\label{Numerical Data Section}

To limit the complexity of our data set, we first focused on the 505 common stocks issued by 500 large-cap companies that traded on American stock exchanges. The S\&P 500 index, which measures the performance of these companies, is one of the most commonly used indices to represent the U.S. stock market. However, the S\&P 500 index is a dynamic composition of the 500 currently most valuable companies. If we just analyze the 505 stocks which are currently on the list, we may then be biased in a sense that we already selected ``winners'', so we also included the historical S\&P 500 companies to mimic a realistic environment. In addition, to ensure the accuracy and reliability of the stock price data, we collected stock data from two different sources on the same dates, namely Yahoo Finance\cite{yahooapi} and Alpha Vantage\cite{alphavantageapi}, and controlled the adjusted closing prices across the two sources to within 2\% error. The remaining stocks after data cleaning include 451 current and 67 past S\&P 500 companies, due to the cases of acquisition, split-up, privatization, bankruptcy, and missing values, etc. The detailed data cleaning steps are in the \hyperref[appendix]{appendix}. For the preprocessing after the data cleaning, we converted the adjusted closing prices into a log return format, which is the difference in the weekly average log adjusted closing price. The relation between the weekly average log adjusted closing price $log(p_t)$ and the log scale return $R_t$ is defined as:

$$R_t = log(p_t) - log(p_{t-1}) $$

The log difference format allows easier comparisons among different stocks, and helps these time series to appear stationary. The final processed data contained stock information from January 1, 2000 to December 31, 2019, with the starting day for different stocks varying, as shown in Table~\ref{summary_of_data}.\\

\begin{minipage}{1\linewidth}\centering
\begin{tabular}{l|c|c|}
\cline{2-3}
                                            & \multicolumn{1}{l|}{S\&P 500 index} & \multicolumn{1}{l|}{S\&P 500 stocks} \\ \hline
\multicolumn{1}{|l|}{\# of observations}    & 1,043                               & 427,255                              \\ \hline
\multicolumn{1}{|l|}{percentage increasing} & 58.19\%                            & 55.03\%                              \\ \hline
\end{tabular}
\captionof{table}{A summary for the weekly return data for the composite S\&P 500 index, as well as 518 individual S\&P 500 stocks, from January 1, 2000 to December 31, 2019.}\label{summary_of_data}
\end{minipage}\\


\subsection{Technical Indicators and Finance Report Data}

\begin{minipage}{1\linewidth}\centering
\captionof{table}{A summary for the technical indicators we used in the model. These indicators are all calculated based on the daily high, low, open, close, dividend, and volume information.}\label{summary_of_technical_indicators}
\vspace{2mm}
\begin{adjustbox}{width=\columnwidth,center}
\begin{tabular}{|l|l|l|}
\hline
Indicator Name & Quick Introduction                                                                                                                                                                                   & Window period     \\ \hline
CCI            & \begin{tabular}[c]{@{}l@{}}The Commodity Channel Index can help to identify price reversals, price extremes\\ and trend strength.\end{tabular}                                                       & 20 days            \\ \hline
MACDH          & \begin{tabular}[c]{@{}l@{}}The Moving Average Convergence Divergence Histogram is the difference between\\ MACD and the MACD signals.\end{tabular}                                                   & 12, 26, and 9 days \\ \hline
RSI            & \begin{tabular}[c]{@{}l@{}}The Relative Strength Index, which measures the average of recent upward movement\\ versus the upward and downward movements combined, in a percentage form.\end{tabular} & 14 days            \\ \hline
KDJ            & The Stochastic Oscillator measures where the close is relative to the low and high.                                                                                                                  & 14, 3, and 3 days  \\ \hline
WR             & \begin{tabular}[c]{@{}l@{}}The Williams \%R index is another index for buy signals by measuring the current \\ price in relation to the past N periods.\end{tabular}                                 & 14 days            \\ \hline
ATR            & \begin{tabular}[c]{@{}l@{}}The Average True Range provides an indicator for the volatility of price. We \\ converted ATR into percentages.\end{tabular}                                              & 14 days            \\ \hline
CMF            & The Chaikin Money Flow provides an indicator related to the trading volume.                                                                                                                          & 20 days            \\ \hline
\end{tabular}
\end{adjustbox}
\end{minipage}\\

Besides the daily adjusted closing price variable, which we used to calculate our dependent variable, return,  we extracted the daily high, low, open, close, dividend, and volume information in each trading day for the cleaned data of the 518 current and past S\&P 500 stocks. We transformed them into composite technical indicators that are calculated based on these variables. After careful research, we added several indices to measure the trend, momentum, volatility and volume of each stock. As Table~\ref{summary_of_technical_indicators} shows:  The CCI and MACD measure the trend, while RSI, KDJ and WR measure the momentum, the ATR measures the volatility, and the CMF measures the volume of the stock. To remove extreme values, we capped each variable after the preprocessing steps we introduced in later sections. Finally, we took the median of each week as the variable value for our weekly data.\\

\begin{minipage}{1\linewidth}\centering
\captionof{table}{A summary for the financial report variables we used in the model. These variables are calculated using the most recent stock price and quarterly report.}\label{summary_of_financial_report}
\vspace{2mm}
\begin{tabular}{|l|l|}
\hline
\multicolumn{1}{|c|}{Variable} & \multicolumn{1}{c|}{Quick Introduction}                                                                                                \\ \hline
PE                             & \begin{tabular}[c]{@{}l@{}}The price to earning ratio is calculated as the share price \\ divided by the earnings per share.\end{tabular} \\ \hline
PB                             & \begin{tabular}[c]{@{}l@{}}The price to book ratio is calculated as the share price \\ divided by the book value per share.\end{tabular}  \\ \hline
PS                             & \begin{tabular}[c]{@{}l@{}}The price to sales ratio is calculated as the share price\\ divided by the revenues per share.\end{tabular}    \\ \hline
\end{tabular}
\end{minipage}\\

In addition to the stock price information, financial accounting information such as Revenue, Gross Margin, Assets and Liability for public companies can be found at the 10-Q quarterly report from the U.S. Securities and Exchange Commission. However, for convenience, we chose to use the financial reports data that Yahoo Finance collects. Also, because the collected financial reports dataset included several hundred variables, as an alternative, we instead utilized three commonly used valuation measurement indices to help to predict the return from the earning, booking value, and revenue perspective, as shown in Table~\ref{summary_of_financial_report}. As with the technical indicators, we capped each variable after preprocessing, and took the median value  of each week as the variable value for our weekly data.\\


\subsection{Text News Data}\label{Text Data Selection}

\begin{minipage}{1\linewidth}\centering
\captionof{table}{An example of NYT text data after selecting the variables for display and modifying the structure}\label{text_keyword_example}
\begin{adjustbox}{width=0.9\columnwidth,center}
\newcolumntype{Y}{>{\small\raggedright\arraybackslash}X}
\begin{tabularx}{140mm}{lYlYl}
\hline
Snippet           & "Chris Cox, who quit Facebook last year after differences with the company’s chief executive, Mark Zuckerberg, is returning as chief product officer." \\   \hline
Web url           & "https://www.nytimes.com/2020/06/11/technology/facebook-chris-cox.html"   \\   \hline
Lead paragraph    & "SAN FRANCISCO — Facebook said on Thursday that Chris Cox, a former top executive, was returning to the company as chief product officer." \\   \hline
Headline           & \begin{tabular}[c]{@{}l@{}} \{main: "Facebook Brings Back a Former Top Lieutenant to Zuckerberg", \\  sub: None \} \end{tabular} \\   \hline
Keywords           & \begin{tabular}[c]{@{}l@{}} {[}\{name: "subject", value: "Social Media", rank: 1 \},\\ \{name: "subject", value: "Mobile Applications", rank: 2 \},\\ \{name: "subject", value: "Appointments and Executive Changes", rank: 3 \},\\ \{name: "organizations", value: "Facebook Inc", rank: 4 \},\\ \{name: "persons", value: "Cox, Chris (1982- )", rank: 5 \},\\ \{name: "persons", value: "Zuckerberg, Mark E", rank: 6 \}{]} \end{tabular} \\ \hline
Other variables & \begin{tabular}[c]{@{}l@{}} \{document type: "article", type of material: "News" , \\ news desk : "Business", section name: "Technology",\\ word count: 370, publish date: "2020-06-11T19:16:33+0000" \}\end{tabular} \\ \hline 
\end{tabularx}
\end{adjustbox}
\end{minipage}\\

Apart from the numerical data, one important feature of our study is the utilization of text data pertaining to news items. There are several newspaper websites such as Financial Times and Thomson Reuters that provide Application Programming Interface (API) access to their database. We chose the New York Times (NYT) Application Programming Interface (API), as for now it provides free access to all historical data, and the data are carefully annotated. The NYT data include the title, the author, the publication time, the type of article, the annotated related organization and person, and the abstract and first paragraph of the article. We selected the period from January 2000 to December 2019, which included 3.2 million collected articles, to align with our numerical datasets. In Table~\ref{text_keyword_example}, we chose several key variables to show an example of the structure of the text data. Also, Figure~\ref{NYT_monthly_summary} shows the NYT article monthly inflow by category for the 20-year period.\\

\begin{minipage}{1\linewidth}\centering
\includegraphics[width=1\linewidth]{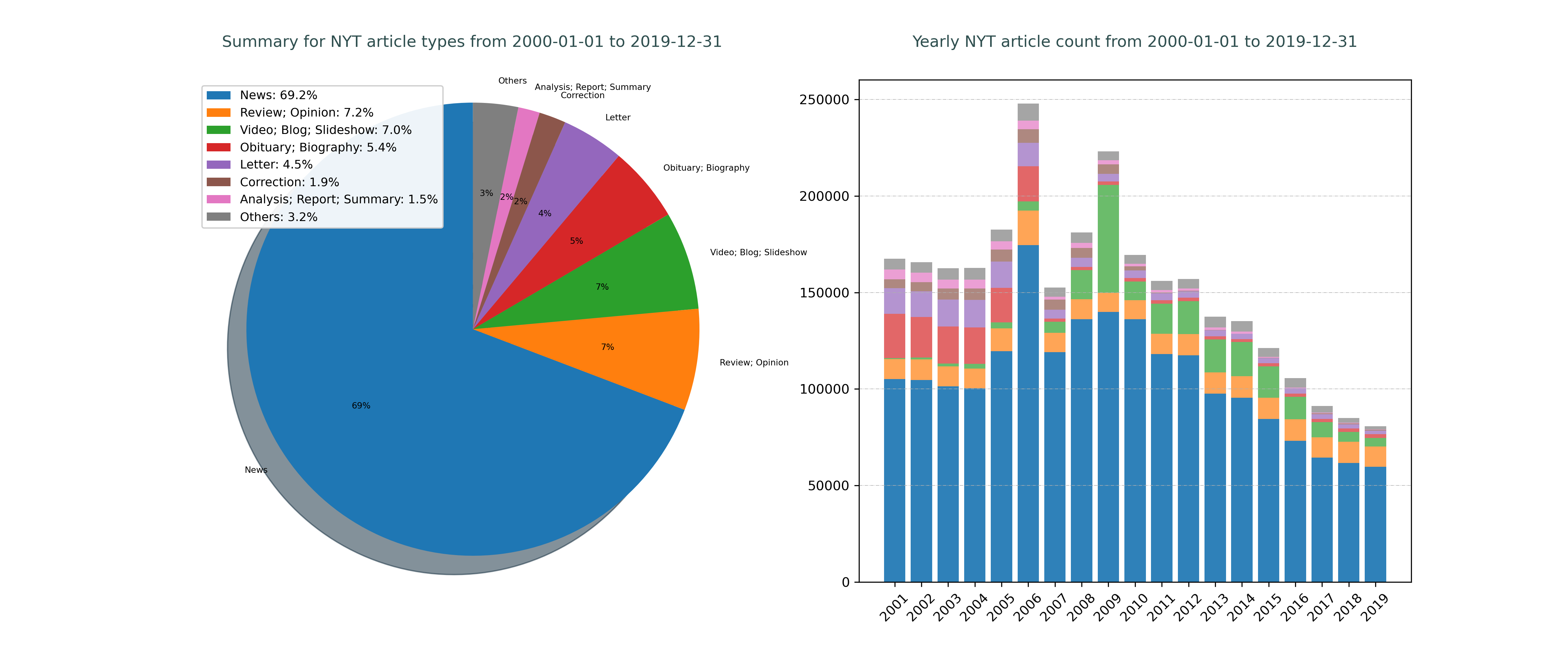}
\captionof{figure}{Summary for NYT news data from January, 2000 to December, 2019.}\label{NYT_monthly_summary}
\end{minipage}

To link these news articles related to each specific stock to the rest of our data, we used the keywords annotated by the NYT database along with text processing techniques to classify which stocks an article refers to. Note that we do not have a direct matching from the S\&P 500 stock tickers to the related keywords mentioned. Furthermore, the company names and the NYT keywords are not in a standard form, having many different expressions for the same thing, while the same entities have different expressions in terms of grammar and semantics. Thus we established a connection between them by first normalizing the NYT keywords and the tickers' company names, then matching them by calculating first edit distance similarity and next cosine similarity of word embeddings.\\

The edit distance is measured by the minimum number of operations needed to convert one string to another. In our preprocessing of strings, we made all characters lowercase (except for capitalizing the first letter of each word), cleaned all symbols and punctuations, and tightened the space between words to exactly one space. Then we set up decision rules using regular expressions to substitute one common word for words of similar meaning. For example, we replaced words like "corporation", "incorporated", and "company" with the "inc" symbol. We included as many replacement rules as possible to account for special cases and common synonyms. We then used the Longest Common Subsequence \cite{hirschberg1977algorithms} metric, which only allows for insertion and deletion operations, and outputs the length of the longest common subsequence after insertion and deletion, to calculate the similarity between the ticker company names and the NYT organization keywords. We scaled the LCS length into percentages relative to the original string lengths for comparison. The portion with highest similarity was then sent to the word embedding steps, as embedding distance calculation takes more computation time.\\

For calculating word embedding \cite{mikolov2013distributed} similarities, we chose to use a pre-trained 128-dimension token-level word embedding with a length-979,661 dictionary, trained on the English Google News 200 Billion corpus, provided by Google. We removed punctuation and split and tightened spaces, following the preprocessing standard of the pre-trained embedding model. For mapping strings into categories in the dictionary, we first checked whether the original string was in the dictionary, then the capitalized word, and next the lowercase word version. Finally for those remaining words, we separated them into pieces that were in the dictionary, by choosing the split with the least amount of splitting. For those words with the same amount of splits, we choose the one that has the lowest standard deviation of the lengths of separated pieces. As all 26 basic letters are in the dictionary, this ensured a mapping with every word in the dictionary. We did not replace words with regular expressions, as the word embedding itself had already learned the semantic similarity. Hereafter, the word level embedding was combined using the Sqrt N metric, which sums the token vectors and then divides by the square root of the number of tokens, to get the embedding for a short phrase and calculate cosine similarity.\\

We generated a list of the highest word embedding matches for each ticker for manual checking and were able to find the linked entities for the ticker companies accurately. A total of 170,779 related news items were obtained for analysis using our method, out of the 3.2 million articles. As expected, the news reports were heavily inclined towards the most renowned companies. Out of the S\&P 500 stocks, there were only 163 that received more than 100 article reports during the 20-year period. Out of these 163 stocks, 36 received more than 1000, as shown in Figure~\ref{NYT_news_count}.\\

\begin{minipage}{1\linewidth}\centering
\includegraphics[width=0.7\linewidth]{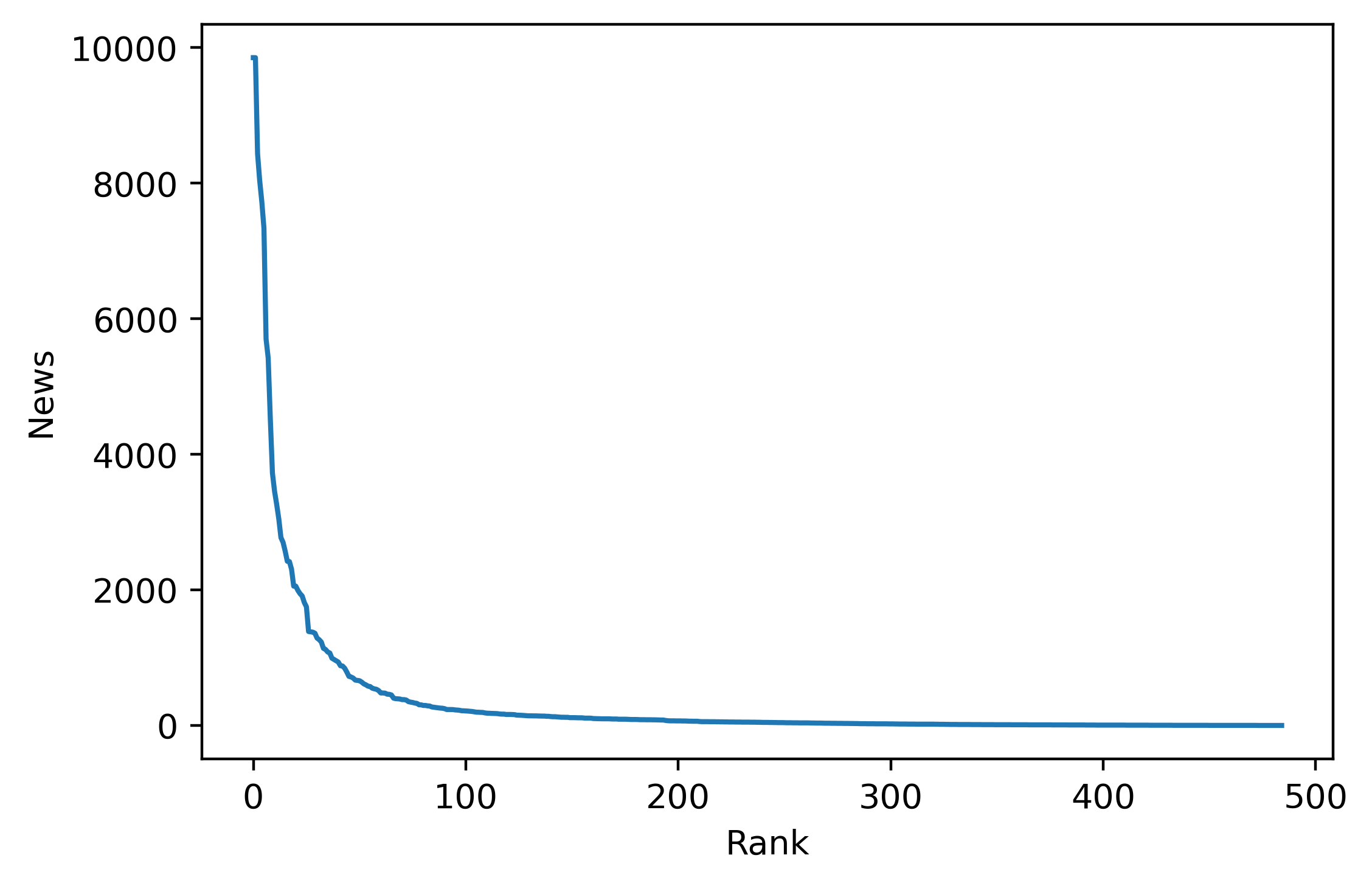}
\captionof{figure}{NYT news count, sorted by the number of news items for each company, from January 2000 to December 2019.}\label{NYT_news_count}
\end{minipage}


\subsection{Sentiment Extraction for Text Data}

With the link between tickers and NYT keyword organizations established, we calculated the sentiment of those related articles. The original BERT-BASE model\cite{devlin2018bert} was obtained by unsupervised learning from a 800 million-word BookCorpus and the 2,500 million-word English Wikipedia dataset, with 12 encoder layers and a hidden size of 768. This BERT-BASE model is already trained by Google and is available publicly. The BERT model with finance domain knowledge\cite{araci2019finbert} was then obtained by further unsupervised tuning of this BERT-BASE model using the Reuters TRC2 finance 29-million-word dataset. Finally, we used labeled finance data from Financial PhraseBank\cite{malo2014good}, which Araci\cite{araci2019finbert} used to test the performance of their model, to train the BERT model to decide the sentiment of each article in the NYT dataset. The Financial PhraseBank data contains 4,840 sentences that were annotated by 16 people with finance backgrounds to judge positive, neutral, or negative tones. We considered subsets of the data for which 100\% and over 66\%, respectively, of people agreed in the judgements, and separated each subset into a training set of 80\% of the observations and a test set of the other 20\%. Using data for which the judgments of all 16 people agreed produced a higher accuracy of 0.97, but we chose to use the more extensive data for which more than 66\% of the 16 people agreed in their judgments which yielded a 0.89 accuracy. As we used the fitted probability of positive opinion minus the probability of negative opinion as the sentiment score, a wider variety of opinions in human judgments would help us in estimating the sentiment.\\

With the model decided, we used the materials from the NYT news database to infer the sentiment. The NYT data contains one main headline, one print headline, a snippet and the first paragraph for each article. We combined them into a big paragraph for the analysis. The combined texts have an average length of 74.48 words with standard deviation of 31.06, with 94.6\% of them being less than 128 words. We set the maximum input length to 128 words, so that sentences longer than 128 words would only have the first 128 words as input, while sentences shorter than 128 words would be padded to 128 words. In our preprocessing, we chose the ``bert-base-uncased'' tokenizer to clean the text. We selected several samples to manually test the performance of the BERT model prediction, as Table~\ref{NYT_BERT_example} shows. Of the 170,779 extracted articles, 23,469 were positive, 37,426 were negative, and 109,884 were neutral. Again, we took the median sentiment each week for each company (with weeks ending on Fridays) as the sentiment variables in our weekly data.\\

\begin{minipage}{\textwidth}
\centering
\captionof{table}{Examples for the BERT model judgment for NYT data.}\label{NYT_BERT_example}
\begin{adjustbox}{width=\columnwidth,center}
\begin{tabular}{|l|c|c|}
\hline
\multicolumn{1}{|c|}{Content concatenated from title, snippet, and part of first paragraph}                                                                                                                                                                                                                                                                                                                                                                                                                                                                                                                                                                                                                                                                                                                                                                                                                                                    & \begin{tabular}{c} Predicted Class\\ Probabilities\end{tabular}                                                                     & sentiment \\ \hline
\begin{tabular}[c]{@{}l@{}}``MARRIOTT TO HALT 35 SENIOR-LIVING COMMUNITIES  Marriott International to\\  cancel up to 35 senior-living communities that are in early stages of developing, citing what\\  it terms 'supply pressures in some markets' (S) Citing what it termed ''supply pressures in \\ some markets,'' Marriott International said yesterday that it would cancel up to 35 senior-living\\  communities that are in early stages of development. The Washington-based hotel company, \\ which currently operates 139 senior-living centers and is building an additional 15, also said \\ slow sales at newly opened centers were expected to reduce earnings by 5 cents a share in the \\ fourth quarter. The projects to be canceled represent about half of the senior-living projects in \\ the company's development pipeline, Marriott said.''\end{tabular} & \begin{tabular}[c]{@{}c@{}}{[}0.0213, positive \\ 0.9707, negative\\ 0.008, neutral{]}\end{tabular} & -0.9494   \\ \hline
\begin{tabular}[c]{@{}l@{}}``Technology Briefing | Hardware: Advanced Micro Ships Faster Chips Advanced Micro\\ Devicees begins selling faster computing Advanced Micro Devices, Intel's top rival in the\\ market for personal computer processors, began selling faster models of its chips. The new\\ versions of Advanced Micro's flagship Athlon chip run at 1.4 gigahertz, while its Durons,\\ for less expensive home PC's, hit 950 megahertz, John Crank, its marketing manager, said. \\ Compaq Computer and Gateway will use the Athlon chips in systems available immediately,\\ he said.''\end{tabular}                                                                                                                                                                                                               & \begin{tabular}[c]{@{}c@{}}{[}0.8819, positive\\ 0.0226, negative\\ 0.0955, neutral{]}\end{tabular}   & 0.8593   \\ \hline
\begin{tabular}[c]{@{}l@{}}``General Motors Chief to Hand Over Reins to Company's President General Motors Corp\\ chairman-chief executive John F Smith Jr says he will hand his chief executive title and\\ day-to-day responsibility for running company to G Richard Wagoner Jr, president and chief\\ operating officer, starting in June; Smith will... John F. Smith Jr., the chairman and chief\\ executive of General Motors announced today that he would hand his chief executive title and\\ day-to-day responsibility for running the company, the world's largest automaker, to G. Richard\\ Wagoner Jr., 46, the president and chief operating officer, starting in June.''\end{tabular}                                                                                                                                                                                                               & \begin{tabular}[c]{@{}c@{}}{[}0.0136, positive\\ 0.0025, negative\\ 0.9839, neutral{]}\end{tabular}   & 0.011   \\ \hline
\end{tabular}
\end{adjustbox}
\end{minipage}\\


\section{Preprocessing, Evaluation Metric and Separation of Training and Test Data}\label{evaluation_section}

The training and test data were separated as follows: We used three years of information as training data. After two years of training data were input, the individual model predicts future observations iteratively, calculates the prediction errors and updates the model at the end of each year based on these errors. Also starting from the third year, the ensemble models were fitted using past years' prediction outputs as inputs to the ensemble model. In this way, we tracked the performance of different models at various regions of the time domain. As a preprocessing step, all predictor variables were transformed to standard Gaussian distributions via Yeo-Johnson transformations \cite{yeo2000new} on the training data, and we filled missing values with 0. In the test set, the absolute values of these preprocessed predictor variables were capped at 4.5 to prevent undue influence by extreme values. The dependent variable was still the return at time $t+1$. Also, we used several metrics to measure the performance based on the true return $R$ and the predicted return $\widehat{R}$:\\ 

Direction prediction accuracy (DA) at time $t$ for the past $n$ periods:

$$DA_{n,t} = \frac{1}{n} \sum_{i=0}^{n-1}{ \mathbbm{1}_{[\mathbbm{1}_{[R_{t-i} \geq 0]} = \mathbbm{1}_{[\widehat{R_{t-i}} \geq 0]}]}}$$

Upward direction prediction accuracy (UDA) at time $t$ for the past $n$ periods:

$$UDA_{n,t} = \begin{cases} 
							1 &, \mathrm{if} \sum_{i=0}^{n-1}{ \mathbbm{1}_{[R_{t-i} \geq 0]} } = 0\\
							\frac{ \sum_{i=0}^{n-1}{ \mathbbm{1}_{[R_{t-i} \geq 0]} \cdot  \mathbbm{1}_{[\widehat{R_{t-i}} \geq 0]}} }{  \sum_{i=0}^{n-1}{ \mathbbm{1}_{[R_{t-i} \geq 0]} } }&, \mathrm{otherwise}
							\end{cases}
$$

Downward direction prediction accuracy (UDA) at time $t$ for the past $n$ periods:

$$DDA_{n,t} = \begin{cases} 
							1 &, \mathrm{if} \sum_{i=0}^{n-1}{ \mathbbm{1}_{[R_{t-i} < 0]} } = 0\\
							\frac{ \sum_{i=0}^{n-1}{ \mathbbm{1}_{[R_{t-i} < 0]} \cdot  \mathbbm{1}_{[\widehat{R_{t-i}} < 0]}} }{  \sum_{i=0}^{n-1}{ \mathbbm{1}_{[R_{t-i} < 0]} } }&, \mathrm{otherwise}
							\end{cases}$$

Root Mean square error (RMSE) at time $t$ for the past $n$ periods:

$$RMSE_{n,t} = \sqrt{\frac{1}{n}\sum_{i=0}^{n-1}{ (R_{t-i} - \widehat{R_{t-i}})^2 } }	 $$

Mean absolute error (MAE) at time $t$ for the past $n$ periods:

$$MAE_{n,t} = \frac{1}{n}\sum_{i=0}^{n-1}{ |R_{t-i} - \widehat{R_{t-i}}| } 	 $$

For each individual stock, for all stocks combined, and for the S\&P 500 index, we calculated these metrics for each year, and for all 20 years of data.\\


\section{Interpretation of Variables}

Before performing predictions for each individual company, we tested the existence of linear relationships among different variables and the dependent variable (return at time $t+1$) for the 427,255 observations from 518 stocks. We explored these relationships across different companies. Figure~\ref{val_to_return_summary} shows scatterplots of the dependent variable return at $t+1$ against various preprocessed predictor variables at time $t$, from January 1, 2000, to December 31, 2019. From the graphs, possible linear relationships were observed for these variables: $\text{return}_t$, sentiment, CCI, RSI, KDJ, WR, MACDH and CMF. The relationships between the dependent variable and the variables PE, PS, PB, ATR do not appear linear.\\

\begin{minipage}{1\linewidth}\centering
\includegraphics[width=0.65\linewidth]{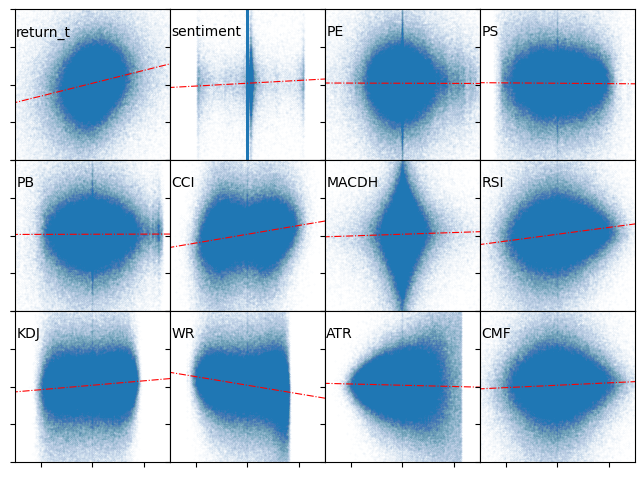}
\captionof{figure}{Relation of each predictor variable at $t$ with the dependent variable return at $t+1$, from January 1, 2000, to December 31, 2019, where these variables were transformed to standard Gaussian and missing values filled by 0. These graphs share a x-axis range from -3 to 3 and y-axis range of -10\% to 10\%.}\label{val_to_return_summary}
\end{minipage}\\

Also, due to the sparsity of news items, 392,333 of the 427,255 of the observed variable pairs have missing values on the variable sentiment score. To check the relation between return and sentiment, we excluded missing values instead of filling them by 0, and plotted the sentiment versus the return at $t+1$ for each year, as Figure~\ref{sentiment_to_return_summary} shows. In general, we expect a positive news sentiment at time $t$ produces a higher return at $t+1$, and the 18 positive slopes out of 20 aligned with our expectation. However, we observe a decreasing trend of the slope parameter over time, which might indicate that as the spread of information has become faster during the 20-year period, the effect of news on next week's stock price has become weaker, as Figure~\ref{sentiment_parameter_value} shows. Note the weekly news data were delimited by each Friday, recognizing the fact that stocks do not trade on weekends.\\

\begin{minipage}{1\linewidth}\centering
\includegraphics[width=0.7\linewidth]{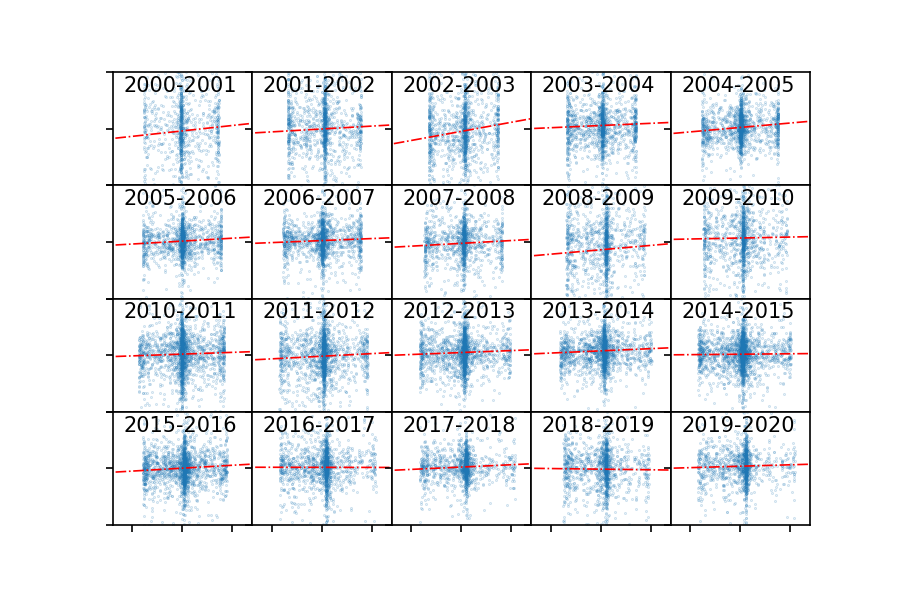}
\captionof{figure}{Scatter plot of sentiment at $t$ on x-axis versus the dependent variable return at $t+1$ on y-axis, with the dashed line representing the simple fitted regression line for the two variables. These plots were generated every year, from January 1, 2000, to December 31, 2019, with missing values excluded. Out of the 20 years, 18 have a positive slope.}\label{sentiment_to_return_summary}
\end{minipage}\\

\begin{minipage}{1\linewidth}\centering
\includegraphics[width=0.5\linewidth]{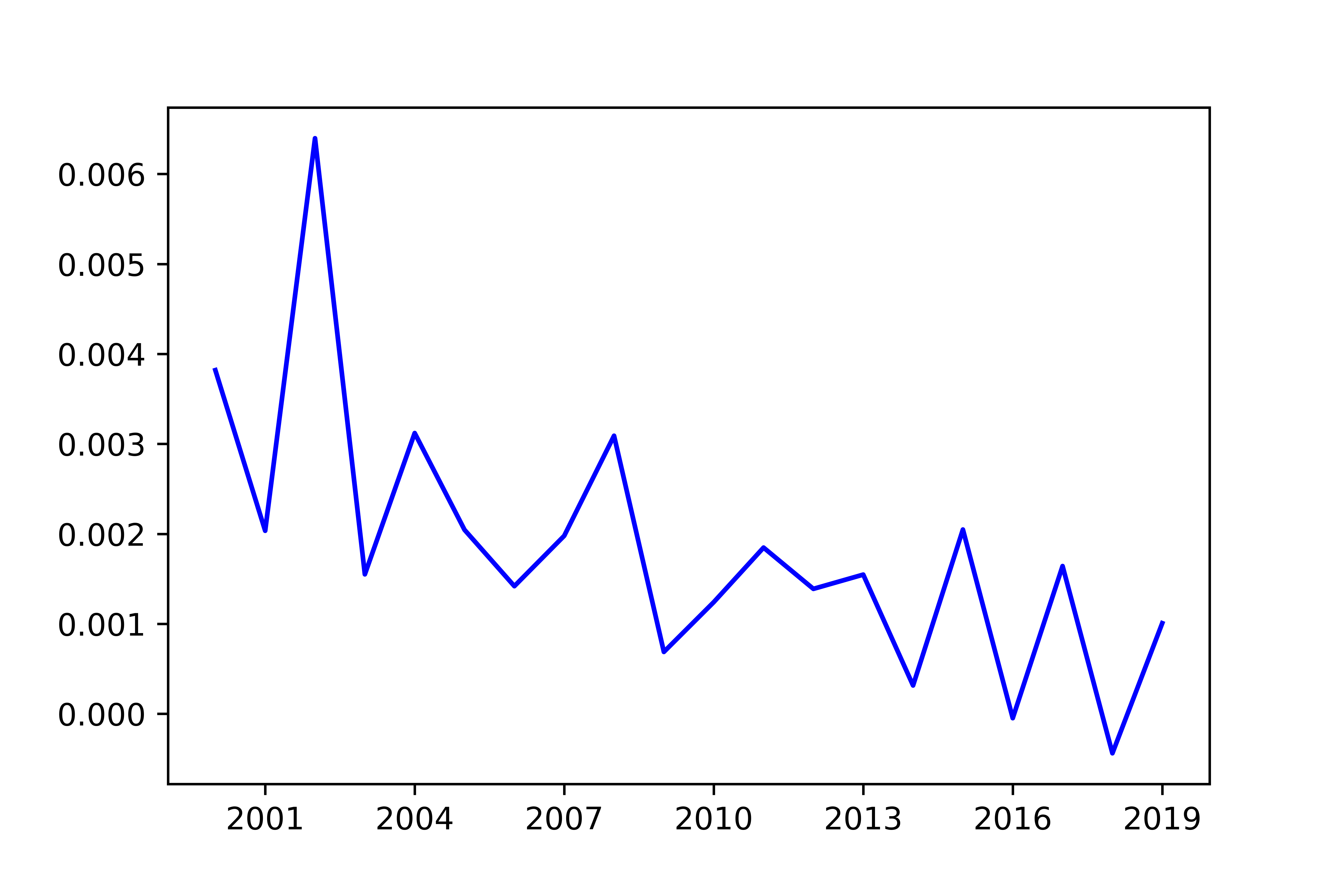}
\captionof{figure}{The parameter values of the 20 simple linear regressions fitted for each year sentiment at $t$ versus the dependent variable return at $t+1$. Out of the 20 years, 18 have a positive slope. However, a decreasing trend in slope over time was also observed.}\label{sentiment_parameter_value}
\end{minipage}\\

For fundamental variables PE, PS and PB, we found that beginning in the year 2000, there was a linear trend showing a lower PE, PS, or PB ratio corresponding to a higher return at $t+1$. However, that linear trend diminished starting in 2003 and became negligible after 2007. We investigated the relationship between PE, PS, and PB and return at $t+1$ among different sectors, for each individual company. Figure~\ref{indicces_by_sector} shows the fitted regression lines for each individual company for return at $t+1$ against PS, separate by the 11 industry sectors. Out of the 515 S\&P 500 companies (3 excluded due to missing values), 86.6\% have a negative slope, which aligns with our intuition that a lower price to sales ratio produces a higher potential return. When only considering year 2007 and hereafter, this percentage drops to 81.6\%. When fitting regression lines of the return at $t+1$ against PE and PB, respectively, 67.7\% and 78.1\% of the slopes are negative, with this percentage varying across sectors.\\ 

\begin{minipage}{1\linewidth}\centering
\includegraphics[width=0.6\linewidth]{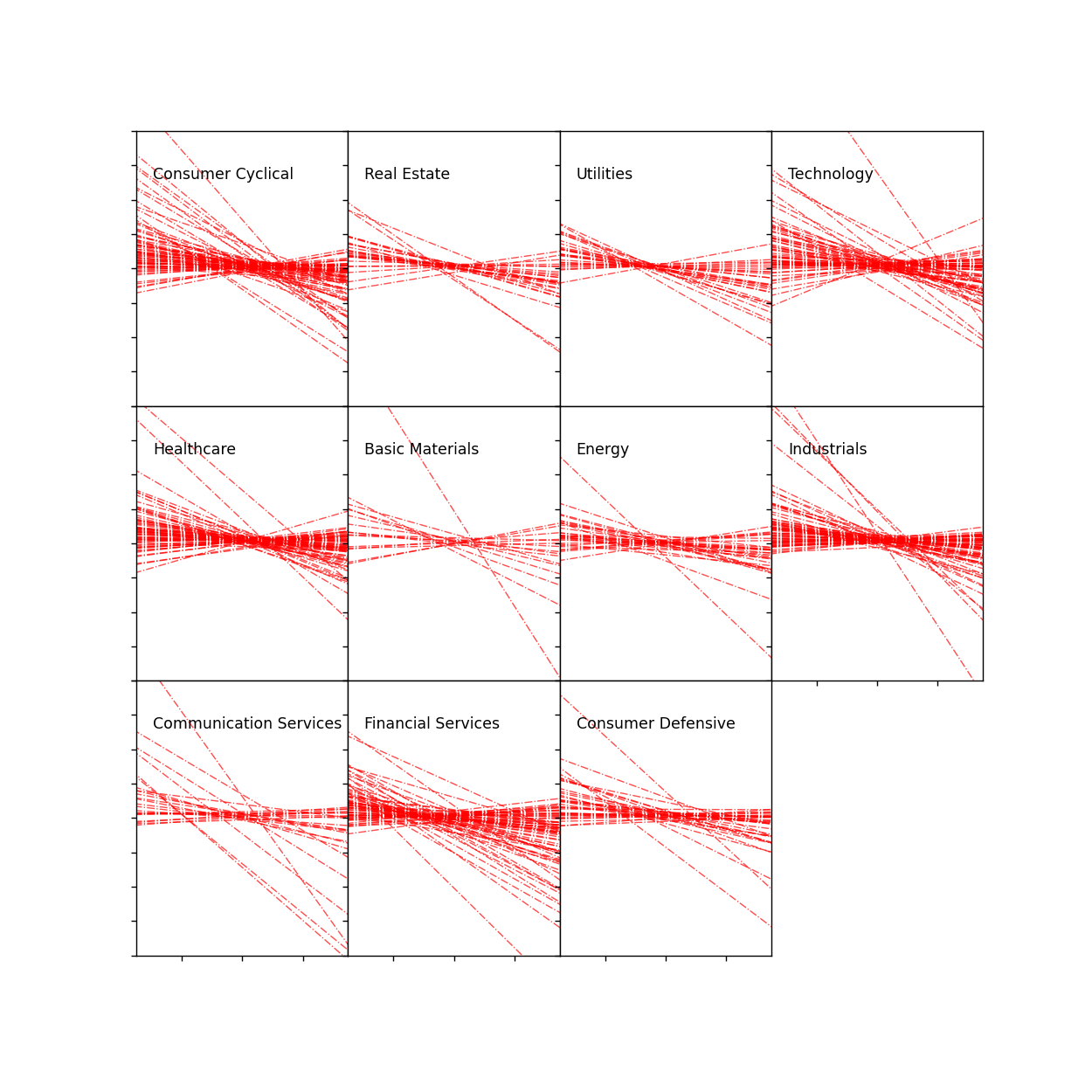}
\captionof{figure}{The fitted regression lines for each individual company for return at $t+1$ against PS, separate by the 11 industry sectors.}\label{indicces_by_sector}
\end{minipage}\\


\section{Applied Models}

Our study of these technical, fundamental, and text-based sentiment variables revealed that different types of variables should enter our model in different forms. Because information on sentiment was sparse, sentiment variables were best utilized in models using combined stock data. We included variables PE, PS, and PB only in models that did not assume a linear functional form. Also, although linear trends were detected with the other variables, the relationships are not totally linear, and variables like ATR have more to do with the volatility than mean trends in stock prices. To solve these somewhat conflicting requirements, we set up ensemble stacking models to combine different individual models together to make predictions. The models we considered in combination were as follows:\\

\subsection{ARIMA Model}

The Autoregressive Integrated Moving Average (ARIMA) model is a linear model that predicts a time-indexed observation using linear combinations of past values and errors\cite{ariyo2014stock}. We performed an automatic search for the best ARIMA model by setting $p$, $q$, $d$ as tuning parameters to optimize: We first determined $d$ by using both the Kwiatkowski-Phillips-Schmidt-Shin (KPSS) test and Augmented Dickey-Fuller test, choosing the larger $d$ calculated by these two tests, but limiting the largest $d$ to be 4. Then we performed an exhaustive search for the best choice of $p$ and $q$ in terms of the lowest Akaike Information Criterion (AIC), setting the max $p$ and $q$ to be 4 to limit the computation time. The ARIMA model response is univariate; we used the past returns $R_{t} \ldots, R_{t-p+1}$ to predict the next return $R_{t+1}$, for each individual company.

\subsection{Linear Regression Model}

The linear regression model has the advantage of being both easy to interpret and easy to extend to a multivariate form. We used a linear model for each individual company, each sector, and for the whole set of S\&P 500 stocks. For each of these models, we used the variables return, CCI, MACDH, RSI, KDJ, WR, CMF and sentiment at time $t$ to predict return at $t+1$.

\subsection{Tree Based Random Forest Model}

The Random Forest\cite{ho1995random} is a model based on decision trees that resolves the problem of overfitting existing in simple decision tree models by bagging on training data and variables: i.e., randomly sampling the training data with replacement and randomly selecting variables that are used to predict during each iteration. We performed a grid search of the number of estimators, maximum depth, and maximum number of variables to find the best fit. With variables return, CCI, MACDH, RSI, KDJ, WR, CMF, PE, PS, and PB, we applied the random forest for each individual company and each sector.

\subsection{Feed Forward Neural Network Model}

The Feed Forward Neural Network (FFNN)\cite{svozil1997introduction} is a neural network structure that links every input from the previous layer to the nodes in the next layer. The variables we used in the FFNN model are return, CCI, MACDH, RSI, KDJ, WR, CMF, ATR, PE, PS, and PB at time $t$. We utilized a model structure with one hidden layer of 48 nodes, relu activation\cite{brownlee2019gentle}, and a dropout \cite{srivastava2014dropout} of 0.6 to construct our model. Each of the nodes in the hidden layer is calculated by linear combinations of the input variable values with the activation function applied, and then the predicted return is calculated by a direct linear combination of the 48 node values in the hidden layer. Also, dropout rate 0.6 means for each node $1,2, \ldots, 48$, input elements are randomly set to zero with 0.6 probability during training to prevent overfitting.

\subsection{Long Short-Term Memory Model}

The Long Short-Term Memory (LSTM) model\cite{gers1999learning} has a flexible functional form that does not require assumptions on the error terms and incorporates temporal structures with a time window: Instead of putting inputs at every time point in the window period together in one function, LSTM uses a summary of previous inputs, along with the current input, to predict the next output recursively. LSTM also has a forgot-gate structure that allows filtering unimportant information and keeping key signals. The variables we used in the LSTM model were return, CCI, MACDH, RSI, KDJ, WR, CMF, ATR, PE, PS, and PB, from the previous three time periods of $t-2$, $t-1$, and $t$. We trained the model on multiple timesteps simultaneously by having input and output sequences of length three and predicting the response one time unit ahead. The LSTM layer we chose had the structure of return sequence, with 32 nodes, Tanh activation function, and dropout rate of 0.6. Similar to the FFNN model, the predicted return was calculated three times recursively by a direct linear combination of 32 LSTM node output values for time points $t-1$, $t$, and $t+1$. The predictions at $t-1$ and $t$ helped training, while the prediction at $t+1$ really predicted the future. In this way, we increased the amount of data available for training. We also adopted two LSTM model structures: The first consisted of two LSTM layers stacked, while the second had just one LSTM layer. The two-layer model was trained using all available stocks, while the one-layer model was first trained using all available stocks, then fine-tuned again using each individual stock's data for that stock's prediction. The fine tuning was controlled by keeping the LSTM layer parameters frozen, while only training and updating the linear combination parameters with a much smaller learning rate, and limited iterations. Although the fine-tuning led to a lower accuracy, the variability it created eventually boosted the ensemble model accuracy, which we discuss in later sections.\\


\subsection{Summary of individual models}

After carefully tuning each individual model's parameters and preprocessing options, we report the best results obtained with each category of models, by directional accuracy, as Table~\ref{Model_report_list} shows. Other models that were also fitted, but not shown in the table were Support Vector Regression, Kernel Ridge Regression, ExtraTree Decision trees, and Gated Recurrent Units, we do not show them because either they have a relatively lower accuracy in stock prediction problems, or there are similar methods shown in the table that have a slightly better performance.\\

\begin{minipage}{1\textwidth}
\begin{adjustbox}{width=\columnwidth,center}
\begin{tabular}{|l|l|l|l|l|l|l|l|}
\hline
Model                                                                            & Input data                                                                                                                              & tuning parameters                                                                                                             & MAE    & RMSE    & DA    & UDA   & DDA   \\ \hline
\begin{tabular}[c]{@{}l@{}}ARIMA, yearly update\\all past data, for each stock\end{tabular}                  & past return                                                                                                                             & \begin{tabular}[c]{@{}l@{}}max p = 4, max q = 4, \\ d by hypothesis test\end{tabular}                                         & 0.0254 & 0.00157 & 57.17 & 68.27 & 43.37 \\ \hline
\begin{tabular}[c]{@{}l@{}}Linear Regression, all stock\\ ten years rolling, yearly update\end{tabular} & \begin{tabular}[c]{@{}l@{}}return, sentiment, CCI, \\ MACDH, RSI, KDJK , \\ WR, CMF from last period\end{tabular}                       & minimize RMSE                                                                                                                 & 0.0251 & 0.00152 & 59.72 & 65.18 & 52.99 \\ \hline
\begin{tabular}[c]{@{}l@{}}Random Forest, for each sector\\ten years rolling, yearly update \end{tabular}         & \begin{tabular}[c]{@{}l@{}}return, sentiment, PE, \\ PS, PB,CCI, MACDH, \\ RSI, KDJ, WR, ATR, \\ CMF from last period\end{tabular}      & \begin{tabular}[c]{@{}l@{}}max depth=8, max features=all,\\ estimators=400, minimize MAE\end{tabular}                           & 0.0244 & 0.00144 & 60.35 & 74.90 & 42.20 \\ \hline
\begin{tabular}[c]{@{}l@{}}Feed Forward Network, all stock\\ten years rolling, monthly update\end{tabular}                                                                               & \begin{tabular}[c]{@{}l@{}}return, sentiment, PE, \\ PS, PB,CCI, MACDH, \\ RSI, KDJ, WR, ATR, \\ CMF from last period\end{tabular}     & \begin{tabular}[c]{@{}l@{}}one forward layer with 48 nodes,\\ relu activation, Dropout 0.6,\\ minimize MAE\end{tabular}       & 0.0243 & 0.00142 & 60.92 & 78.53 & 38.95 \\ \hline
\begin{tabular}[c]{@{}l@{}}LSTM, all stocks together,\\ all past data, yearly update\end{tabular}             & \begin{tabular}[c]{@{}l@{}}return, sentiment, PE, PS,\\ PB,CCI, MACDH, RSI, \\ KDJ, WR, ATR, CMF\\ from past three periods\end{tabular} & \begin{tabular}[c]{@{}l@{}}2 LSTM layer, 32 nodes each, \\ Dropout 0.6, minimize MAE,\\ 3 outputs with one shift\end{tabular} & 0.0240 & 0.00140 & 61.69 & 74.07 & 46.13 \\ \hline
\end{tabular}
\end{adjustbox}
\captionof{table}{Combined performance of several prediction models on all 518 current and past S\&P 500 stocks, during the period January 1, 2003 to December 31, 2019, with each model updated and tested yearly.}\label{Model_report_list}
\end{minipage}\\

In general, the old traditional ARIMA model performs worse than other models, and we view it as a baseline model. On the other hand, the traditional Linear Regression models trained together using all stock's data have an 2.55\% DA accuracy higher on average than ARIMA, and we treated them as a baseline for multivariate models. Experimental results showed that the best linear regression models trained (1) separately on each individual stock; (2) separately on each sector; and (3) together using all available stocks data combined yield, respectively, 58.16\%, 59.55\%, and 59.72\% overall DA accuracy, as Table~\ref{Model_linear_individual_vs_combine} shows. This trend of accuracy increasing as the data scope grows was observed on other types of models as well, indicating that although different stocks might behave differently, they could be used in the same model with different variable values.\\

\begin{minipage}{1\textwidth}\centering
\begin{tabular}{|l|l|}
\hline
\multicolumn{1}{|c|}{Model}                                                                                     & \multicolumn{1}{c|}{DA} \\ \hline
\begin{tabular}[c]{@{}l@{}}linear regression models trained \\ separately on each individual stock\end{tabular} & 58.16\%                 \\ \hline
\begin{tabular}[c]{@{}l@{}}linear regression models trained \\ separately on each sector\end{tabular}           & 59.55\%                 \\ \hline
\begin{tabular}[c]{@{}l@{}}linear regression models using \\ all stocks data combined\end{tabular}              & 59.72\%                 \\ \hline
\end{tabular}
\captionof{table}{Overall directional accuracy comparison of linear models trained on different levels.}\label{Model_linear_individual_vs_combine}
\end{minipage}\\

Using the linear model, we also examined different rolling window periods and found having a rolling period of ten years yields the highest accuracy, compared with using all past data. We then applied the ten-year period again to the Random Forest and FFNN model. We trained the Random Forest model on the subsets by sector level, since training the Random Forest using data from all stocks took much more computation time, and we wanted to have more variability in our ensemble steps. The FFNN and LSTM model were trained by having 10\% of the training data randomly chosen as validation sets for early stopping \cite{prechelt1998early}. We set the FFNN model to be updated monthly as it took less computational time compared to other machine learning methods. In summary, machine learning models, especially the LSTM, had better performance than traditional statistical models. The LSTM model outperformed other individual models in every aspect; one possible reason is that it incorporated an input period from the previous three weeks and thus contained more information. However, in Jiang et al.\cite{jiang2020improved} the Random Forest model showed a higher acurracy by 3.99\% compared with the LSTM. We think this difference in results came from the difference in sample size: our LSTM model had 14,113 parameters for the two-layer stacked model and 5,793 for the one-layer model, compared with our 427,255 observations. In contrast, Jiang et al. had a 4,054 sample but had a three-layer LSTM structure with even more parameters needed to fit it. The combined observations from 518 different stocks helped train these machine learning models, which usually work better on larger datasets. Figure~\ref{model_performance} shows the average directional accuracy by year, for different models.\\

\begin{minipage}{1\linewidth}\centering
\includegraphics[width=0.6\linewidth]{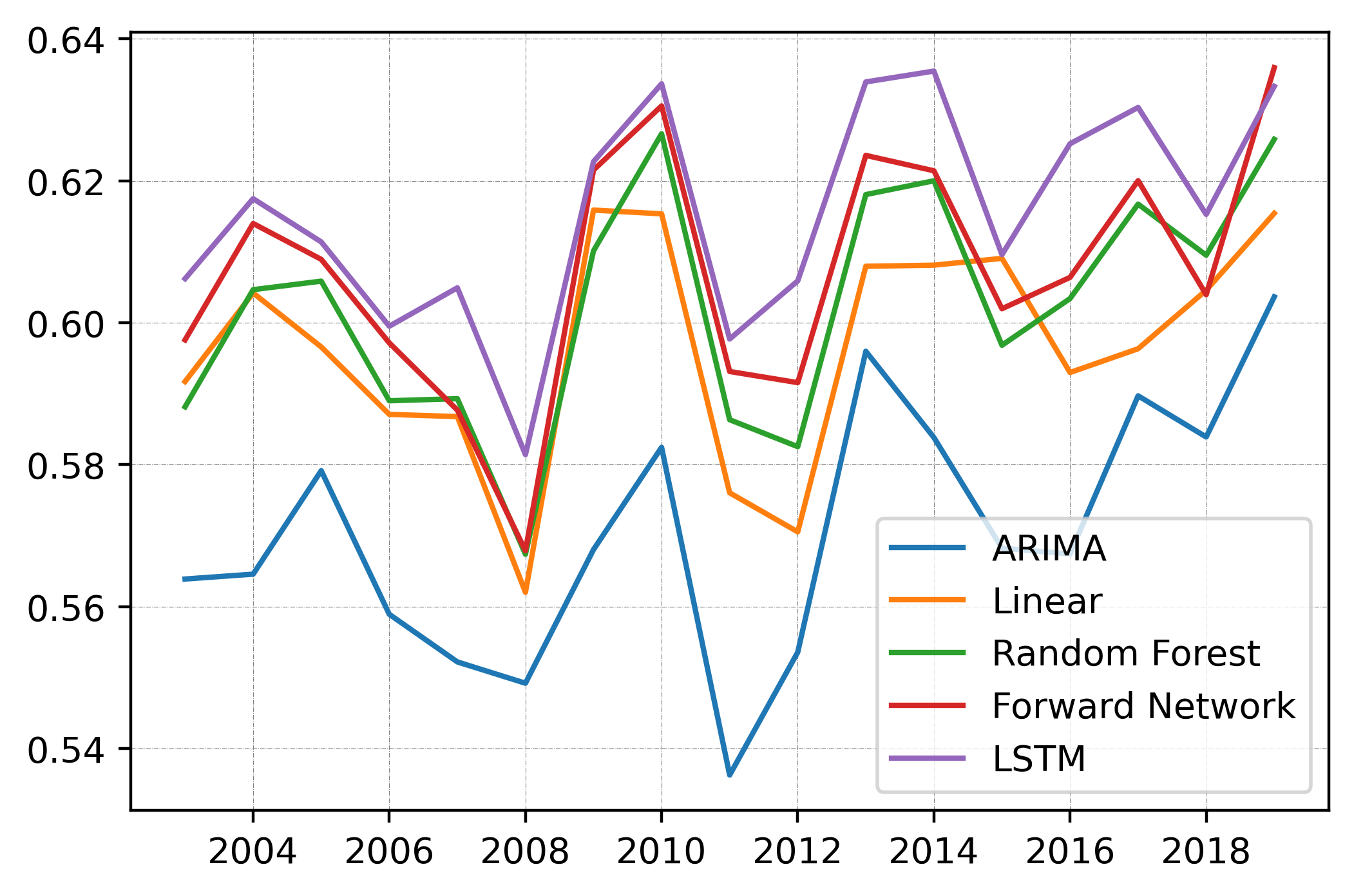}
\captionof{figure}{The average directional accuracy of different models by year.}\label{model_performance}
\end{minipage}\\


\section{Combining Models with Ensemble Stacking}

\subsection{Ensemble Stacking}

After we obtained predictions from the individual models with different input variables, we combined these prediction results into an ensemble model. One natural way of combining different models is for these individual models to vote for whether the stock price will go up or down in the next period. The vote can either be a majority vote, or a weighted vote. Another natural way of combining different models is through choosing the best model for each year or for each stock, based on the past performance: although selecting models by cross validation generally cannot exceed the performance of the best model, the prediction for all S\&P 500 stocks can be separated into different years and stocks, which when combined yields a better overall performance.  Based on these two natural approaches, we utilized an ensemble stacking model to determine the weight of each individual model each year. Ensemble stacking \cite{wolpert1992stacked} is a technique that uses several base models' predicted outputs as predictor variables for the final prediction. In our study, we used a linear regression as the meta-learner to learn the weight of each base model, forcing the constant term in the regression to be zero, and all weights (coefficients) to be nonnegative during the combination stage for each year.\\

\subsection{Summary of Ensemble Models}

Recall that our ensemble model was fitted using past years' prediction outputs as inputs, starting from the third year. The ensemble model we chose was based on a combination of four machine learning models: Random Forest model fitted for each sector, FFNN fitted monthly, two-layer stacked LSTM, and the LSTM model fine-tuned on each individual stock. Figure~\ref{ensemble_model_performance} shows the yearly directional accuracy comparison for the ensemble model versus the best individual model for that year. Table~\ref{ensemble_report_list} shows the detailed structure for the four individual models we chose for the ensemble, as well as the performance of the combined ensemble model built based on their outputs. We can see the ensemble model outperforms every individual model in terms of MAE, RMSE, and DA. The LSTM model further tuned using each stock's data boosted the ensemble model with a 0.3\% increase in directional accuracy. Likewise, both the Random Forest and FFNN model contributed to a higher directional accuracy for the ensemble model, by a smaller margin. Adding other models, such as the Gated Recurrent Units (GRU) model into the ensemble, yielded still higher directional accuracy, but the improvement was minimal and GRU is very similar in structure to LSTM, so we preferred the simpler ensemble of four individual models.\\

\begin{minipage}{1\linewidth}\centering
\includegraphics[width=0.8\linewidth]{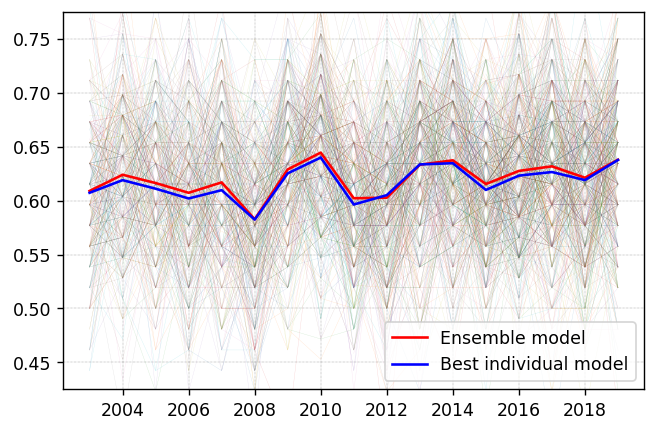}
\captionof{figure}{The yearly DA comparison between ensemble model and the best individual model for that year. The thin lines represent the ensemble DA of different individual stocks by year.}\label{ensemble_model_performance}
\end{minipage}\\

\begin{minipage}{1\textwidth}
\begin{adjustbox}{width=\columnwidth,center}
\begin{tabular}{|l|c|l|l|l|l|l|l|}
\hline
Model                                                                                    & \multicolumn{1}{l|}{Input data}                                                                                                                          & Tuning parameters                                                                                                             & MAE    & RMSE    & DA    & UDA   & DDA   \\ \hline
\begin{tabular}[c]{@{}l@{}}Random Forest, for each sector\\ten years rolling, yearly update \end{tabular}                 & \multirow{2}{*}{\begin{tabular}[c]{@{}c@{}}return, sentiment, PE, \\ PS, PB,CCI, MACDH, \\ RSI, KDJ, WR, ATR, \\ CMF from last period\end{tabular}}      & \begin{tabular}[c]{@{}l@{}}max depth=8, max features=all,\\ estimators=400, minimize MAE\end{tabular}                           & 0.0244 & 0.00144 & 60.35 & 74.90 & 42.20 \\ \cline{1-1} \cline{3-8} 
\begin{tabular}[c]{@{}l@{}}Feed Forward Network, all stocks\\ten years rolling, monthly update\end{tabular}                        &                                                                                                                                                          & \begin{tabular}[c]{@{}l@{}}one forward layer with 48 nodes,\\ relu activation, Dropout 0.6,\\ minimize MAE\end{tabular}       & 0.0243 & 0.00142 & 60.92 & 78.53 & 38.95 \\ \hline
\begin{tabular}[c]{@{}l@{}}LSTM with all stocks together, fine\\ tuning on each stock, all past data\end{tabular} & \multirow{2}{*}{\begin{tabular}[c]{@{}c@{}}return, sentiment, PE, PS,\\ PB,CCI, MACDH, RSI, \\ KDJ, WR, ATR, CMF\\ from past three periods\end{tabular}} & \begin{tabular}[c]{@{}l@{}}1 LSTM layer, 32 nodes each, \\ Dropout 0.6, minimize MAE,\\ 3 outputs with one shift\end{tabular} & 0.0245 & 0.00146 & 61.07 & 67.75 & 52.61 \\ \cline{1-1} \cline{3-8} 
\begin{tabular}[c]{@{}l@{}}LSTM, all stocks together,\\ all past data, yearly update\end{tabular}                      &                                                                                                                                                          & \begin{tabular}[c]{@{}l@{}}2 LSTM layer, 32 nodes each, \\ Dropout 0.6, minimize MAE,\\ 3 outputs with one shift\end{tabular} & 0.0240 & 0.00140 & 61.69 & 74.07 & 46.13 \\ \hline
\begin{tabular}[c]{@{}l@{}}Ensemble, yearly update,\\ two year rolling\end{tabular}                                                           & \begin{tabular}[c]{@{}l@{}}Individual Model outputs\\ from above four model\end{tabular}                                                                                                                                                    & \begin{tabular}[c]{@{}l@{}}Regression with constant fixed \\ at 0, coefficient forced to be \\ greater or equal than 0\end{tabular} & \textbf{0.0239} & \textbf{0.00138} & \textbf{62.09} & 73.71 & 47.61 \\ \hline
\end{tabular}
\end{adjustbox}
\captionof{table}{Combined performance of several prediction models, as well our ensemble model, on all 518 current and past S\&P 500 stocks during the period January 1, 2003 to December 31, 2019.}\label{ensemble_report_list}
\end{minipage}\\

Through repeated experimentation on different combinations of models, we found that forcing the intercept to be 0 was important for our ensemble stacking model to perform well. Holding other conditions unchanged, the ensemble model with no restriction on the intercept would have a 0.67\% lower directional accuracy, causing the ensemble model to perform worse than the individual LSTM model. Also, if an individual model had a much lower directional accuracy than other models, including it in the ensemble would hinder the ensemble model performance. When the chosen individual models performed similarly well, a larger variability in terms of MAE or RMSE among the individual models helped the ensemble model to perform better. This need for variability led us to make different combinations of input data and models in our ensemble model. Also, we investigated the period needed to determine the individual model weights for the final ensemble model, and a grid search ranging from one month to several years revealed that a rolling period of two years and an update period of one year yields the best performance, with all choices of rolling period and update periods yielding a better result than the best individual model. Figure~\ref{model_weight} summarizes the change across years in the weights placed on different individual models, for our ensemble model that predicts 518 S\&P 500 stocks.\\

\begin{minipage}{1\linewidth}\centering
\includegraphics[width=0.6\linewidth]{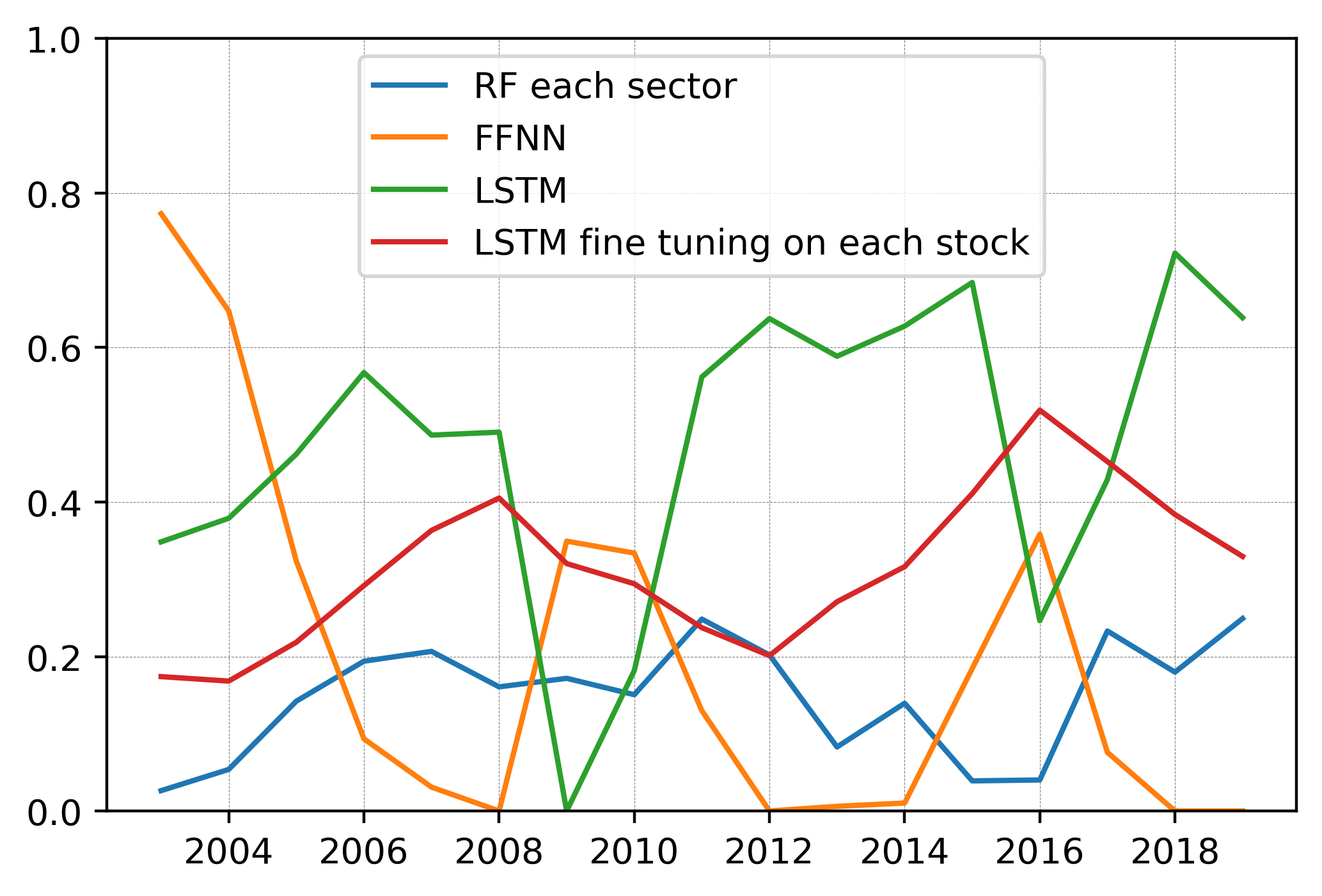}
\captionof{figure}{The weights placed on different individual models for the ensemble model predicting 518 S\&P stocks, by year.}\label{model_weight}
\end{minipage}\\

Besides making predictions for each individual stock, we also built the ensemble model for predicting the S\&P 500 index. We found that compared with models using only the past S\&P 500 index data, the ensemble model built using all combined data coming from the 518 individual stocks had an incredible increase in predicting directional accuracy for the S\&P 500 index. We used the medians of the 518 outputs from the previously shown linear regression, Random Forest, FFNN, and two LSTM models on the individual stock data as inputs for our ensemble model for the S\&P 500 index, and chose a one-year rolling period. Table~\ref{sp500_index_compare_list} shows the models that only used the past S\&P 500 index data, as well as the ensemble model performance in DA accuracy.\\

\begin{minipage}{1\textwidth}
\begin{adjustbox}{width=\columnwidth,center}
\begin{tabular}{|l|l|l|c|}
\hline
\multicolumn{1}{|c|}{Model} & \multicolumn{1}{c|}{Input data}                                                                                                                                                            & \multicolumn{1}{c|}{Tuning parameter}                                                                                                                        & DA                                    \\ \hline
Linear Regression           & \begin{tabular}[c]{@{}l@{}}return, CCI, MACDH, RSI, KDJK , WR, CMF of the \\ S\&P 500 index from last period\end{tabular}                                                                  & minimize RMSE                                                                                                                                                & 58.62\%                               \\ \hline
Random Forest               & \begin{tabular}[c]{@{}l@{}}return, CCI, MACDH, RSI, KDJ, WR, ATR, CMF\\ of the S\&P 500 index, median PE, PB, PS from individual stocks\\ and the sentiment of news about S\&P 500 index  from last period\end{tabular} & \begin{tabular}[c]{@{}l@{}}max depth=8, max features=8,\\ estimators=4000, minimize MAE\end{tabular}                                                         & 60.20\%                               \\ \hline
Ensemble model              & \begin{tabular}[c]{@{}l@{}}median of the 518 outputs from the previously \\ shown linear regression, Random Forest, FFNN, \\ and two LSTM models for individual stocks\end{tabular}        & \begin{tabular}[c]{@{}l@{}}Regression with constant fixed at 0, \\ coefficient forced to be greater or \\ equal than 0, rolling period one year\end{tabular} & \multicolumn{1}{l|}{\textbf{66.18\%}} \\ \hline
\end{tabular}
\end{adjustbox}
\captionof{table}{Comparison of the directional accuracy for different models predicting the S\&P 500 index, during the period January 1, 2003 to December 31, 2019. Because of the small size of the training data set, the FFNN and LSTM models trained using only past index data had unstable results with DA ranging from 55\% to 61\%.}\label{sp500_index_compare_list}
\end{minipage}\\

Note that stock prices have a general long-term tendency to increase over time. As a simple tool to assess accuracy, consider comparing proposed prediction models based on their improvements in directional accuracy over a hypothetical model that always predicts increases for stocks. Table~\ref{model_compare_list} shows such an accuracy comparison between our model and the models proposed by the four authors introduced in the review literature. Compared with the 1.82\% improvement by Jaing et al., 3.48\% improvement by Yu et al., 4.53\% improvement by Ding et al., and 4.09\% improvement by Novak et al., our model has a larger 5.53\% improvement for the DA of S\&P 500 index, and 6.63\% improvement for the DA of individual stocks, compared to a model that only predicts increases.\\

\begin{minipage}{1\textwidth}
\begin{adjustbox}{width=\columnwidth,center}
\begin{tabular}{|l|l|l|c|c|c|}
\hline
Author                                                                     & \begin{tabular}[c]{@{}l@{}}model refresh and\\ predicting period\end{tabular}           & Predicting                                                      & \multicolumn{1}{l|}{\begin{tabular}[c]{@{}l@{}}Predicted \\ accuracy\end{tabular}} & \multicolumn{1}{l|}{\begin{tabular}[c]{@{}l@{}}Percentage of time stock\\ price increased in test set\end{tabular}} & \multicolumn{1}{l|}{\begin{tabular}[c]{@{}l@{}}Improved accuracy compared\\ to model only predicting upwards\end{tabular}} \\ \hline
\begin{tabular}[c]{@{}l@{}}Jiang et al.\cite{jiang2020improved} for \\ S\&P 500 index\end{tabular} & \begin{tabular}[c]{@{}l@{}}no refresh, \\ 09/01/2012-04/01/2019\end{tabular}            & \begin{tabular}[c]{@{}l@{}}closing on\\ next month\end{tabular} & 69.17\%                                                                            & 67.35\%                                                                                                                & 1.82\%                                                                                                                    \\ \hline
\begin{tabular}[c]{@{}l@{}}Yu an Yan\cite{yu2020stock} for\\ S\&P 500 index\end{tabular}      & \begin{tabular}[c]{@{}l@{}}yearly update,\\ 01/01/2010-12/29/2017\end{tabular}          & \begin{tabular}[c]{@{}l@{}}closing on\\ next day\end{tabular}   & 58.07\%                                                                            & 54.59\%                                                                                                                & 3.48\%                                                                                                                    \\ \hline
\begin{tabular}[c]{@{}l@{}}Ding et al.\cite{ding2015deep} for\\ S\&P 500 index\end{tabular}    & \begin{tabular}[c]{@{}l@{}}no refresh,  \\ 02/22/2013- 11/21/2013\end{tabular}          & \begin{tabular}[c]{@{}l@{}}closing on\\ next day\end{tabular}   & 64.21\%                                                                            & 59.68\%                                                                                                                & 4.53\%                                                                                                                    \\ \hline
\begin{tabular}[c]{@{}l@{}}Gorenc Novak and Velušček\cite{gorenc2016prediction}\\ for 370 S\&P stocks\end{tabular}  & \begin{tabular}[c]{@{}l@{}}20 days update,\\ 10/27/2005-06/14/2013\end{tabular}         & \begin{tabular}[c]{@{}l@{}}high on\\ next day\end{tabular}      & 61.16\%                                                                            & 57.07\%                                                                                                                & 4.09\%                                                                                                                    \\ \hline
\begin{tabular}[c]{@{}l@{}}Our model for\\ S\&P 500 index\end{tabular}     & \begin{tabular}[c]{@{}l@{}}monthly \& yearly mixed\\ 01/01/2003-12/31/2019\end{tabular} & \begin{tabular}[c]{@{}l@{}}closing on\\ next week\end{tabular}  & 66.18\%                                                                            & 60.65\%                                                                                                                & \textbf{5.53\%}                                                                                                           \\ \hline
\begin{tabular}[c]{@{}l@{}}Our model for\\ 518 S\&P stocks\end{tabular}    & \begin{tabular}[c]{@{}l@{}}monthly \& yearly mixed\\ 01/01/2003-12/31/2019\end{tabular} & \begin{tabular}[c]{@{}l@{}}closing on\\ next week\end{tabular}  & 62.12\%                                                                            & 55.49\%                                                                                                                & \textbf{6.63\%}                                                                                                           \\ \hline
\end{tabular}
\end{adjustbox}
\captionof{table}{Comparison of the improvement in DA among different models. Our model for the S\&P 500 index was fitted using the same ensemble method but with the median of the individual stock model outputs.}\label{model_compare_list}
\end{minipage}\\

Although a 6.59\% improvement in DA does not seem like a big improvement, in reality, we do not really need to make buy and sell decisions for every stock at every time point. Given realistic conditions, the ensemble model could help make investment decisions: 5.52\% of the time the ensemble model predicted a 2\% or above return in the next period, and given this prediction, 76.52\% of the time the stock price went up in the next period, with an average real return of 2.80\%. 6.08\% of the time there existed a 5\% loss in the next period, and given this condition, 66.42\% of the time the ensemble model predicted the direction right, with an average predicted loss of 0.83\%. We also found in our model that the past DA performance for specific stocks has little impact on the future DA accuracy for that stock. Figure~\ref{model_5_percent} showed the spread of the real return from the individual S\&P 500 companies, through January 1, 2003 to December 31, 2019, given the ensemble model predicted an over 5\% return. We can see that the investment opportunities our model predicted exist among different years and different companies.\\

\begin{minipage}{1\linewidth}\centering
\includegraphics[width=0.8\linewidth]{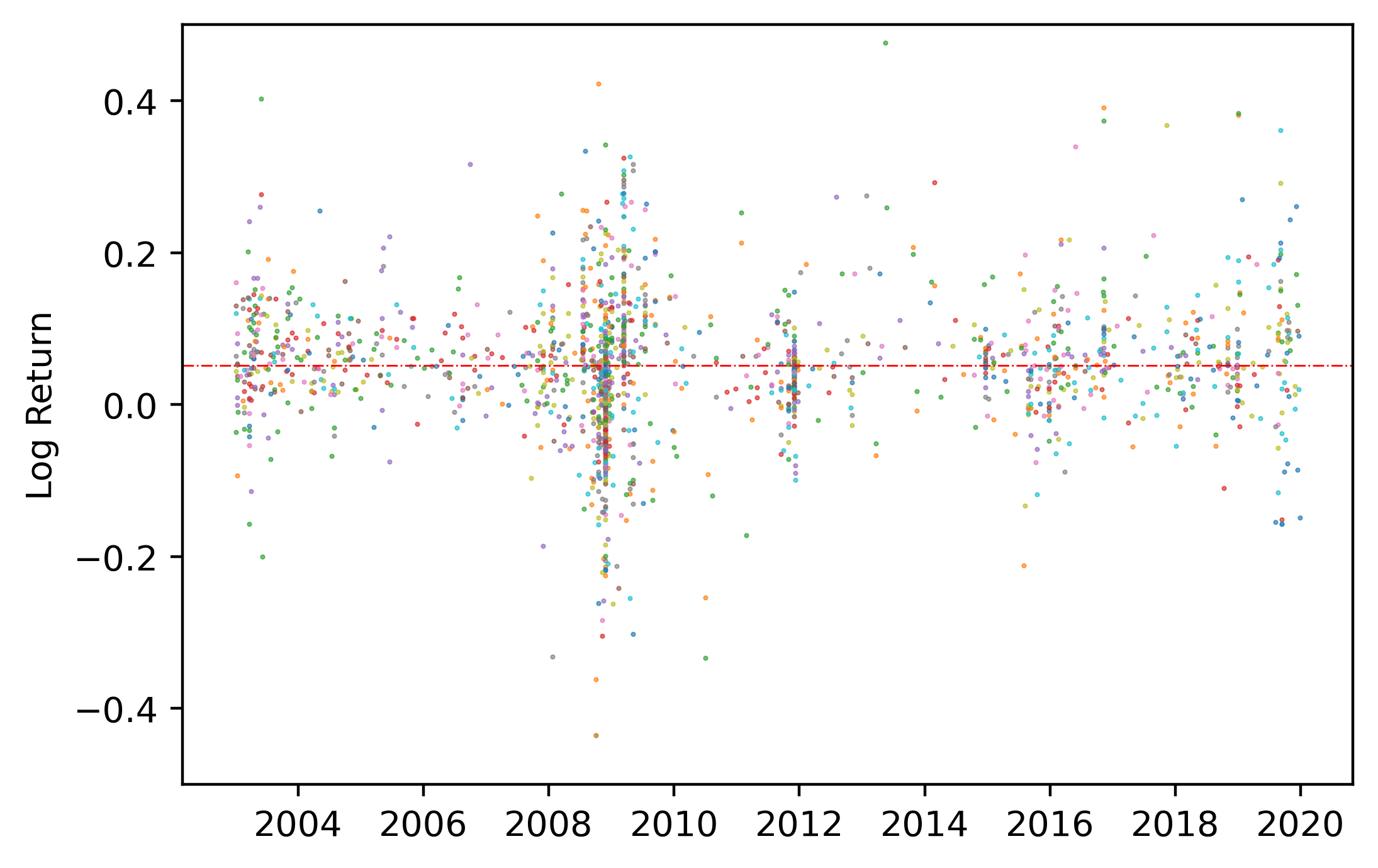}
\captionof{figure}{The spread of time when the ensemble model predicted an over 5\% return, with y-axis the real return. Different colors indicate different stocks, and the red dashed line indicates the average real return of 5.15\%, given the condition.}\label{model_5_percent}
\end{minipage}\\


\section{Conclusion}

We summarized several different approaches to predict stock price, and then integrated those methods together to predict the S\&P 500 stock data. Compared with similar works which apply their methods to a smaller set of data, our data set included text from news items, technical indicators, and fundamental reports for 518 current and former S\&P 500 large-cap companies over a 20-year period to test and train our algorithm, which relates more closely to the real world market. Also, we integrated new machine learning methods such as the Random Forest and LSTM models in an ensemble setting to predict future stock prices with a higher accuracy and adaptability using our combined model compared to uni-method models.\\

Future research may extend beyond stock predicting:  Another important aspect of stock trading is decision making. When outcomes of decisions are partly random, we can model them through the Markov Decision Process (MDP) framework with Reinforcement Learning (RL) techniques \cite{sutton2018reinforcement}\cite{puterman2014markov}.  This is a self-taught learning method to solve MDP problems when the environment is partly unobservable \cite{deng2016deep}. In the stock trading problem, if we model agent states as the currently held stocks and at each week we buy, sell, or hold, then this can be modeled in the MDP framework to achieve goals such as maximizing the long-term return while keeping the risk as low as possible at the same time.\\

\bibliographystyle{plain}
\bibliography{refs}

\section{Appendix: Data Cleaning Steps}\label{appendix}

We actually collected the data on September 30, 2020, so the stocks we collected for our research were stocks listed in S\&P 500 index as of September 30, 2020, plus the stocks that had been removed from the list but were still trackable. From January 1, 2000 to September 30, 2020, there were 247 times in which a common stock was once on the S\&P 500 list but was removed due to reasons like market capitalization changes, merger and acquisition, or being spun off, etc. We summarize the reasoning for removal, as well as whether we decided to track the stock, in Figure~\ref{sp_stock_removel_reason}. Here we justify the reasoning: 122 stocks were removed from the S\&P 500 list due to being merged or acquired (M\&A). Since the stock stops being traded in the stock exchange once the company associated with the stock gets acquired (it will be turned into cash or the shares of the purchasing company, depending on the buying terms), we decided not to track these companies. 102 stocks were removed from the S\&P 500 list due to market reasons such as market value decline. Among these, we tracked 61 of the stocks. For the remaining 41 stocks, eventually 17 went bankrupt, 18 were acquired or merged, one was split, and five had too many missing values or insufficient data. For similar reasons, we tracked six additional former S\&P 500 stocks from the other category, accounting our 67 former S\&P 500 stocks.\\

\begin{minipage}{1\linewidth}\centering
\includegraphics[width=0.6\linewidth]{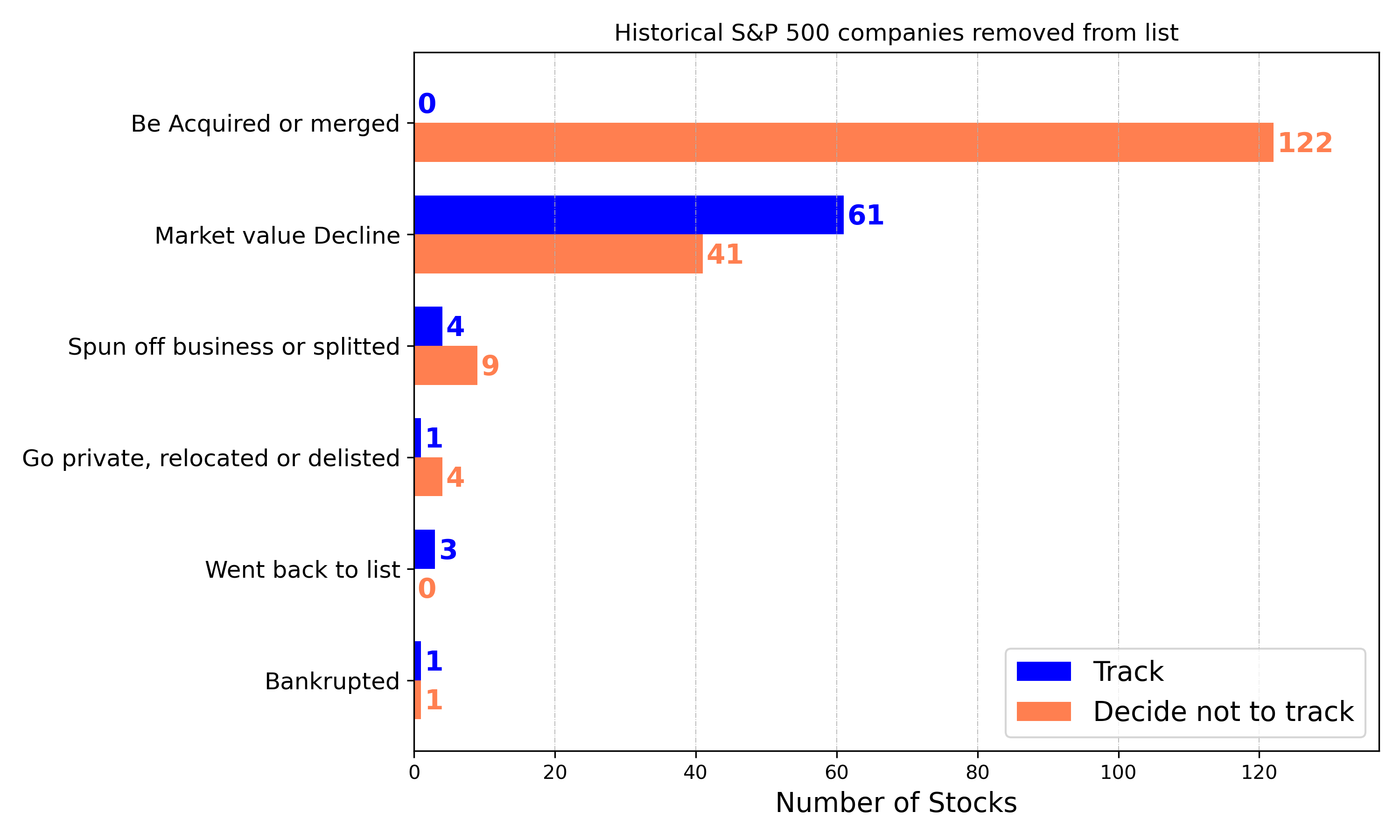}
\captionof{figure}{Summary of the reasons for removal from S\&P 500 list, at the time the common stock was removed, for historical S\&P 500 stocks, for the period from January 1, 2000 to September 30, 2020.}\label{sp_stock_removel_reason}
\end{minipage}\\

For the missing values, we utilized the two data sources \cite{yahooapi} and \cite{alphavantageapi}, and matched them by dates. Considering both current and past S\&P 500 stocks, 69 have missing values. 47 out of them have just one day of data missing, while the other 22 stocks have an average of 2507 days missing. We first deleted missing dates and only considered dates both sources had data, and with a maximum of 10 consecutive days of gaps allowed. Then we deleted any questionable data after matching adjusted closing prices across the two sources to within 2\% error, by choosing the \cite{yahooapi} data as the base. Table~\ref{table_closing_price_summary} and Table~\ref{table_dividend_split_total_summary} show a summary for the error rates. After these steps, stocks that had less than 5 years of trading history after filtering were then also removed, since models training on individual stocks need sufficient data.\\

\begin{minipage}{1\linewidth}\centering
\captionof{table}{Report on the reliability of stock price data, accounted for all the stocks we tracked, from November 1, 1999 to September 30, 2020, from the two different sources \cite{yahooapi} and \cite{alphavantageapi}.}\label{table_closing_price_summary}
\begin{tabular}{|r|c|c|c|c|c|c|}
\hline
\multicolumn{1}{|l|}{}          & \textbf{Open} & \textbf{High} & \textbf{Low} & \textbf{Close} & \textbf{Adj Close} & \textbf{Volume} \\ \hline
\multicolumn{1}{|c|}{\begin{tabular}[c]{@{}c@{}} Percentage of data points\\ within 1\% error \end{tabular}} & 97.07\%       & 97.53\%       & 97.53\%      & 97.54\%        & 77.11\%            & 73.41\%         \\ \hline
\multicolumn{1}{|c|}{\begin{tabular}[c]{@{}c@{}} Percentage of stocks have \\ 99\% quantile error\\ less than 5\% \end{tabular}} & 96.01\%            & 96.01\%  & 96.01\%  & 96.01\%  & 71.06\%  & 40.03\%             \\ \hline
\end{tabular}
\end{minipage}

\begin{minipage}{1\linewidth}\centering
\captionof{table}{Report on the reliability of dividends and splits data, from \cite{yahooapi} and \cite{alphavantageapi}. The amount exactly matched and amount within 1\% error are conditioned on dates being matched.}\label{table_dividend_split_total_summary}
\begin{tabular}{|l|l|l|}
\hline
                        & Dividend & Split \\ \hline
Percentage of dates matched   &   98.08\%       &  76.15\%     \\ \hline
Amount exactly matched&   88.30\%       & 96.30\%    \\ \hline
Amount within 1\% error &   91.98\%       & 98.83\%    \\ \hline
\end{tabular}
\end{minipage}\\

\end{document}